\documentclass[letterpaper, 10 pt, conference]{ieeeconf} 

\IEEEoverridecommandlockouts 

\usepackage{times} 
\usepackage{amsmath} 
\usepackage{amssymb} 
\usepackage{graphicx}
\usepackage{cases}
\usepackage{algorithm}
\usepackage{algorithmic}
\usepackage{caption}
\usepackage{color}
\usepackage{subfigure}
\usepackage{cite}

\title{\LARGE \bf
Deep reinforcement learning of event-triggered communication and control for multi-agent cooperative transport
}

\author{Kazuki Shibata$^{1,2}$, Tomohiko Jimbo$^{1}$ and Takamitsu Matsubara$^{2}$
\thanks{$^{1}$The authors are with the Autonomous Distributed Cooperative Control Program, Data Analytics Research-Domain, Toyota Central R$\&$D Labs., Inc., 41-1, Yokomichi, Nagakute, Aichi, Japan.
{\tt\small kshibata@mosk.tytlabs.co.jp}}%
\thanks{$^{2}$The authors are with Graduate School of Science and Technology, Division of Information Science, Nara Institute of Science and Technology, Nara, Japan
{}}%
}

\begin{document}

\maketitle
\thispagestyle{empty}
\pagestyle{empty}

\begin{abstract}
In this paper, we explore a multi-agent reinforcement learning approach to address the design problem of communication and control strategies for multi-agent cooperative transport. 
Typical end-to-end deep neural network policies may be insufficient for covering communication and control; these methods cannot decide the timing of communication and can only work with fixed-rate communications. 
Therefore, our framework exploits {\it event-triggered architecture}, namely, a feedback controller that computes the communication input and a triggering mechanism that determines when the input has to be updated again. 
Such event-triggered control policies are efficiently optimized using a multi-agent deep deterministic policy gradient. We confirmed that our approach could balance the transport performance and communication savings through numerical simulations.
\end{abstract}
\section{INTRODUCTION}
Cooperative transport is an important task in multi-agent systems, with applications in the fields of delivery services, factory logistics, and search and rescue.
In particular, to transport large or heavy payloads, the utilization of multi-agent systems has advantages over a single agent, including scalability, flexibility, and robustness to individual agent failures.

\begin{figure}[!tp]
\centering
\subfigure[Event-triggered communication and reconfiguration]{
\includegraphics[width=6cm]{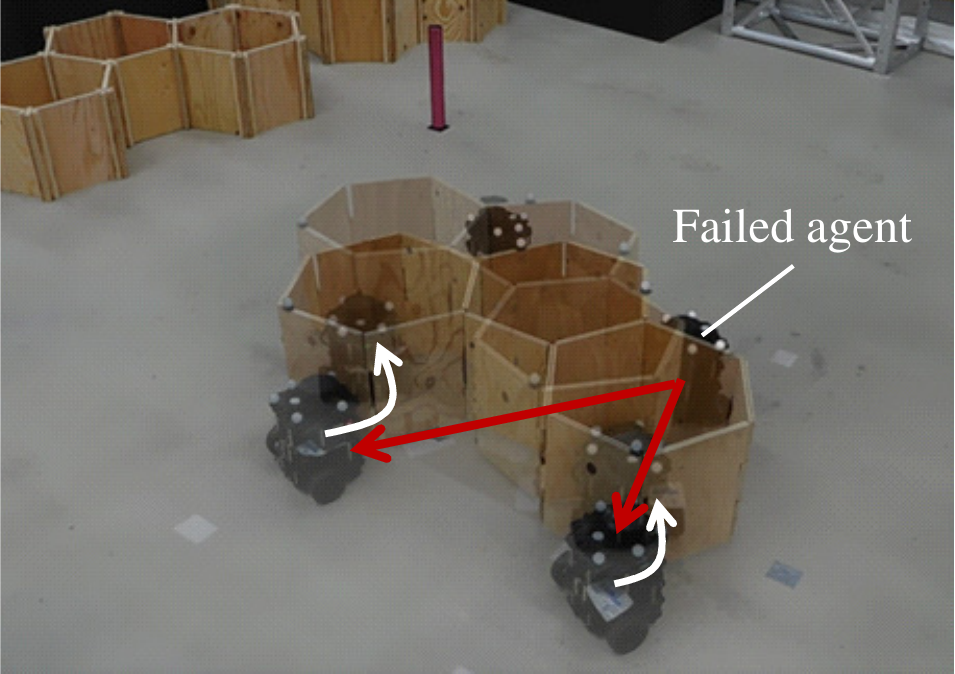}
\label{transport}}
\subfigure[Overview of our framework]{
\includegraphics[width=8cm]{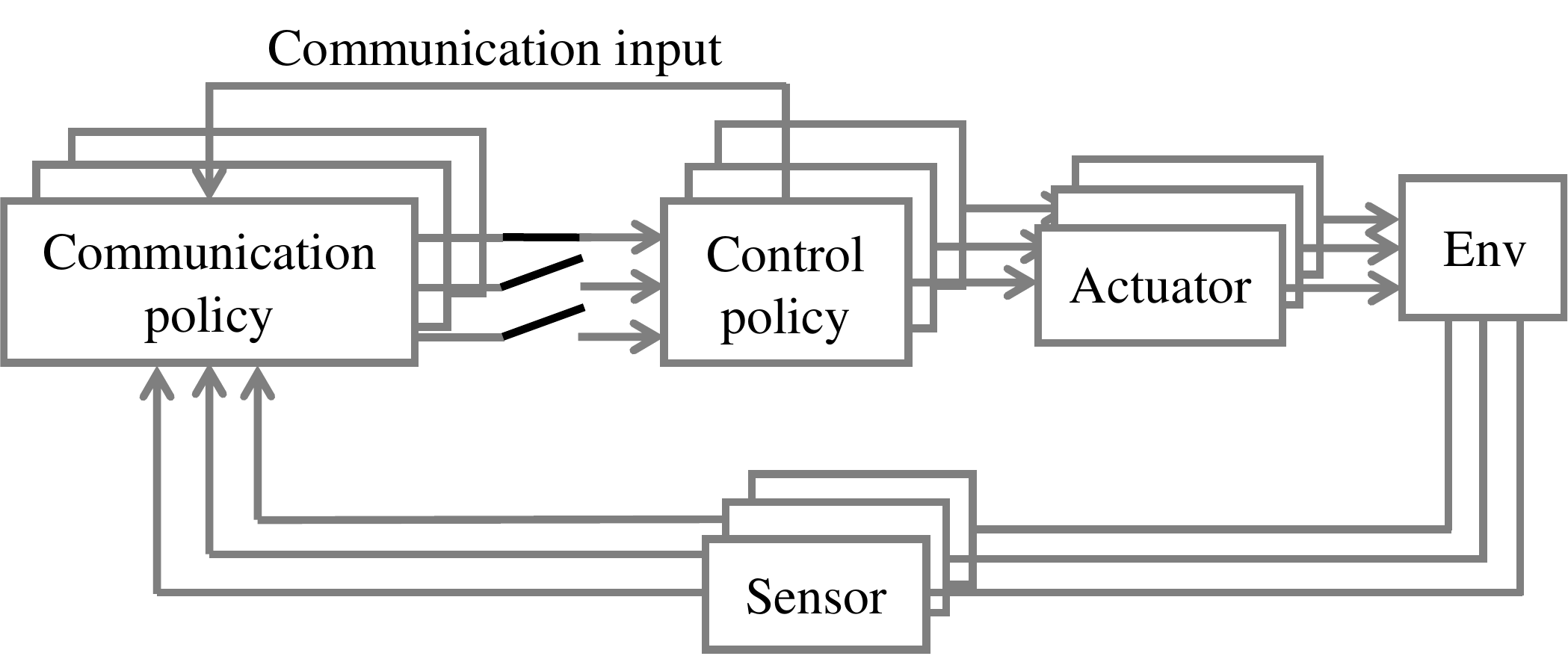}
\label{framework}}
\caption{(a) Event-triggered communication and reconfiguration when one of the agents stops. The red and white arrows represent the direction of communication and the trajectories of the agents, respectively. (b) Overview of our framework.}
\end{figure}

In this paper, we consider cooperative transport using multi-agents, as shown in Fig. \ref{transport}.
Most cooperative transport efforts using multi-agents \cite{Michael2011, Kennedy2015, Loianno2018} have relied on wireless communication to share observations.
In this scenario, they suffer from performance degradation due to packet loss and communication delays as the number of agents and data increases.
Moreover, in scenarios with battery-driven communication, the lifetime can be reduced if each agent communicates with a high frequency.
Therefore, it is crucial to minimize communication.
Although one could derive suitable communication topologies for simple tasks or tasks with strong assumptions \cite{Franchi2014, Franchi2015, Petitti2016, Culbertson2018}, there are few principled approaches to derive minimum-communication strategies in general setups. 
This may be because the minimum-communication strategy's design problem cannot be separated from the design problem of the control strategy for multi-agent systems. Therefore, these problems need to be addressed simultaneously \cite{Baumann2018, Funk}.

In this paper, we explore a multi-agent reinforcement learning (MARL) approach \cite{Lowe2017} to address the simultaneous design problem of communication and control strategies for multi-agent cooperative transport. 
To this end, typical end-to-end deep neural network (DNN) policies may be insufficient for covering communications and control; these methods cannot decide the timing of communication and can only work with fixed-rate communications. 
In contrast, our framework exploits {\it event-triggered architecture} \cite{Dohmann2020}, namely, a feedback controller that computes the communication input and a triggering mechanism that determines when the input has to be updated again, as depicted in Fig. \ref{framework}.
Our framework is inspired by a previous work \cite{Baumann2018} in which the policy model is used to reduce the control signals from a networked controller of a single agent to its actuator.
We extended their policy model in multi-agent setups to reduce the frequency, the number of communicating agents and data.

To confirm the effectiveness of our approach, we conducted two cooperative transport simulations.
We confirmed the effectiveness of our framework through cooperative transport with agents rigidly attached to a payload whose minimum-communication topology could be derived analytically \cite{Culbertson2018}.
Moreover, we confirmed the versatility of our framework through cooperative pushing with actuator failure, whose minimum-communication topology is analytically intractable.

The contributions of the present study can be summarized as follows:
\begin{itemize}
\item We propose a learning framework using event-triggered communication and control in multi-agent setups that balances the transport performance and communication savings.
\item We confirm the effectiveness and versatility of our framework through cooperative transport simulations.
\item We demonstrate our framework in a cooperative transport experiment using multiple ground robots.
\end{itemize}

The remainder of this paper is organized as follows. 
Section II introduces related work. Section III describes the cooperative transport problem with multi-agent communication.
Section IV introduces the mathematical formula for the MARL setting and policy model of event-triggered communication and control. Section V shows the effectiveness of our algorithm through numerical simulations.
Section VI shows the demonstration of the multi-agent policies learned in the simulations. Section VII discusses future work and the limitations of the present study.
Finally, we conclude the paper in Section VIII.

\section{Related Work}
Previous studies on cooperative manipulation using multi-agents have addressed various kinds of tasks, including cooperative pushing \cite{Fink2007, Fink2008}, cooperative manipulation using multiple arm robots \cite{Wang2015, WangICRA2016, WangIJRR2016}, humans and robots \cite{Sieber2015, Gienger2018}. 
Furthermore, several recent studies focused on aerial manipulation by multiple quadcopters using cables \cite{Michael2011, Jiang2013, Sreenath2013} or electromagnetic grippers \cite{Loianno2018, Mellinger2013}.
While these approaches have succeeded in real robot experiments, they have a significant drawback of requiring accurate payload properties, including mass, moment of inertia, or center of gravity.

To solve this problem, Franchi et al. \cite{Franchi2014, Franchi2015} proposed a decentralized estimation and robust control \cite{Petitti2016} in the presence of estimation uncertainties.
However, these algorithms required frequent communication with neighboring agents to perform the estimation.
Without the need for communication, Culbertson et al. \cite{Culbertson2018} proposed distributed adaptive control using the current and desired states of the payload broadcasted to all agents.
They also proved that their algorithm could make the payload state asymptotically converge to the desired state using a Lyapunov function.
In these studies, they derived suitable communication topologies for simple tasks or tasks with strong assumptions.
However, there are few principled approaches to derive minimum-communication strategies in general setups. 

To address this problem, the authors in \cite{Dohmann2020} proposed a novel approach with event-triggered architecture, by which each agent could minimize the frequency of receiving positions and velocities of end effectors from neighboring agents.
Through manipulation experiments using dual-arm robots, they confirmed that their strategy could reduce the communication rate drastically while achieving several manipulation tasks. However, their method requires a dynamics model to design a controller.

Without the need for dynamics models, several authors have proposed DRL approaches to learn communication and control policies simultaneously \cite{Baumann2018, Funk}.
Their policy models are used to reduce the control signals from a networked controller of a single agent to its actuator while performing several control tasks.
In this study, we extended the policy model \cite{Baumann2018} in multi-agent setups to reduce the frequency, the number of communicating agents, and data.

\section{Cooperative Transport Problem with Multi-Agent Communication}

Let us consider a cooperative transport problem using a team of $N$ agents in a 2D obstacle-free environment.
The position, yaw angle, velocity, angular velocity, and desired position of the payload in world coordinates are represented by $\textit{\textbf{x}}\in \mathbb{R}^2$, $\theta \in \mathbb{R}$, $\textit{\textbf{v}}\in \mathbb{R}^2$, $\omega\in \mathbb{R}$, and $\textit{\textbf{x}}^{\ast} \in \mathbb{R}^2$, respectively.

The position, yaw angle, velocity, angular velocity and control input of agent $i\in \mathcal{R}:=\{ 1,\cdots,N\}$ are represented by $\textit{\textbf{x}}_i\in \mathbb{R}^2$, $\theta_i \in \mathbb{R}$, $\textit{\textbf{v}}_i\in \mathbb{R}^2$, $\omega_i\in \mathbb{R}$, $\textit{\textbf{u}}_i\in \mathbb{R}^2$, respectively. Moreover, observations of agent $j\in \mathcal{R}$ held by agent $i$ and observations of all agents held by agent $i$ are represented by $\textit{\textbf{o}}_i^j:=[o_i^{j1},\cdots o_i^{jL}]^{\rm{T}}\it{}\in\mathbb{R}^{L}$ and $\mathcal{O}_i:=[{\textit{\textbf{o}}_i^{1}}^{\rm{T}}\it{},\cdots,{\textit{\textbf{o}}_i^{N}}^{\rm{T}}]^{\rm{T}}\in \mathbb{R}^{N\times L}$, respectively.

The preconditions of this problem are as follows:
\begin{itemize}
\item the mass and the moment of inertia of the payload are unknown;
\item each agent $i$ can determine the agents, the data and timing to communicate.
\end{itemize}

The aim of this study is to control the payload to its desired position while reducing the number of communicating agents and data at each control time step.

According to \cite{Culbertson2018}, we made the following assumptions:
\begin{itemize}
\item all agents know $\textit{\textbf{x}}^{\ast}$;
\item each agent $i$ can observe $\textit{\textbf{x}}$, $\textit{\textbf{x}}_i$, $\theta$, and $\theta_i$.
\end{itemize}

\section{Deep Reinforcement Learning of Event-Triggered Communication and Control}

In this section, we describe the simultaneous learning of control and communication for cooperative transport with multi-agent communication. 
We start with a brief setting of MARL and then introduce our learning framework.

\subsection{Setting of MARL for Cooperative Transport with Multi-Agent Communication}
Let us consider $N$ learning agents whose purpose is to learn the optimal control and communication policies by trial and error through interaction with the environment.
The interaction is mathematically formulated as a Markov decision process.

We denote the sets of states, actions, and observations of $N$ agents as $\mathcal{S}=[\textit{\textbf{s}}_1,\cdots,\textit{\textbf{s}}_N]$, $\mathcal{A}=[\textit{\textbf{a}}_1,\cdots,\textit{\textbf{a}}_N]$, and $\mathcal{O}=[\mathcal{O}_1,\cdots,\mathcal{O}_N]$, respectively.
Each agent $i$ selects action $\textit{\textbf{a}}_i$, under current observations $\mathcal{O}_i$, according to a policy $\pi_i(\textit{\textbf{a}}_i\mid \mathcal{O}_i)$.
Action $\textit{\textbf{a}}_i$ includes the control input as well as variables to determine the agents to communicate and receive data, as described later.
After $N$ agents select the current actions $\mathcal{A}$, the current state $\mathcal{S}$ transitions to the next state $\mathcal{S}'=[\textit{\textbf{s}}'_1,\cdots,\textit{\textbf{s}}'_N]$.
Meanwhile, each agent $i$ receives a reward $r_i$, which is defined by the transport performance with a communication penalty.
Each agent $i$ aims to maximize its own total reward $R_i=\sum^T_{k=1}\gamma^{k-1}r_i(k)$, where $\gamma$ ($0<\gamma\le1$) is a discount factor.

\subsection{Learning Framework of Event-Triggered Communication and Control}
We propose a learning framework for event-triggered communication and control for multi-agent cooperative transport.
Our framework adopted an event-triggered communication architecture and joint policy of communication and control, as shown in Fig. \ref{overview}.

In the event-triggered communication architecture, each agent determines the agents and the data to receive at every control step, as shown in Fig. \ref{ETClaw}.
In the joint policy, each agent calculates the control input at the current control step and the communication input for the next control step.

\subsection{Policy Model}
In \cite{Baumann2018}, the authors proposed an event-triggered policy model to learn the timing to send control signals from a networked controller of a single agent to its actuator. We extend this policy model to multi-agent communication and control.

Let us define the binary variables $w_{ij}$ and $z_{il}$, which represent whether agent $i$ communicates with agent $j$ and whether agent $i$ receives $o_j^{jl}$.
By introducing these variables into the event-triggered law in \cite{Baumann2018}, the event-triggered communication is given by
\begin{eqnarray}
w_{ij}=
\begin{cases}
1, \ \rm{if \ } \> \it{}{c_{ij}}(\mathcal{O}_i)>\rm{0}\\
0, \ \rm{otherwise}\\
\end{cases}
(j=1,\cdots,N),
\label{ETCwho}
\end{eqnarray}
\begin{eqnarray}
z_{il}=
\begin{cases}
1, \ \rm{if \ } \> \it{}{d_{il}}(\mathcal{O}_i)>\rm{0}\\
0, \ \rm{otherwise}\\
\end{cases}
(l=1,\cdots,L),
\label{ETCwhat}
\end{eqnarray}
where $c_{ij}\in \mathbb{R}$ and $d_{il}\in \mathbb{R}$ represent the continuous variables calculated by the joint policy.

Using the event-triggered communication, (\ref{ETCwho}) and (\ref{ETCwhat}), the element of $\mathcal{O}_i$ is updated by
\begin{eqnarray}
o^{jl}_i\leftarrow
\begin{cases}
o^{jl}_j, \ \rm{if \ } \> \it{}w_{ij}z_{il}=\rm{1}\\
-1, \ \rm{otherwise}\\
\end{cases}.
\label{update}
\end{eqnarray}

\begin{figure}[!tp]
\begin{center}
\subfigure[Overview]{
\includegraphics[width=7.5cm]{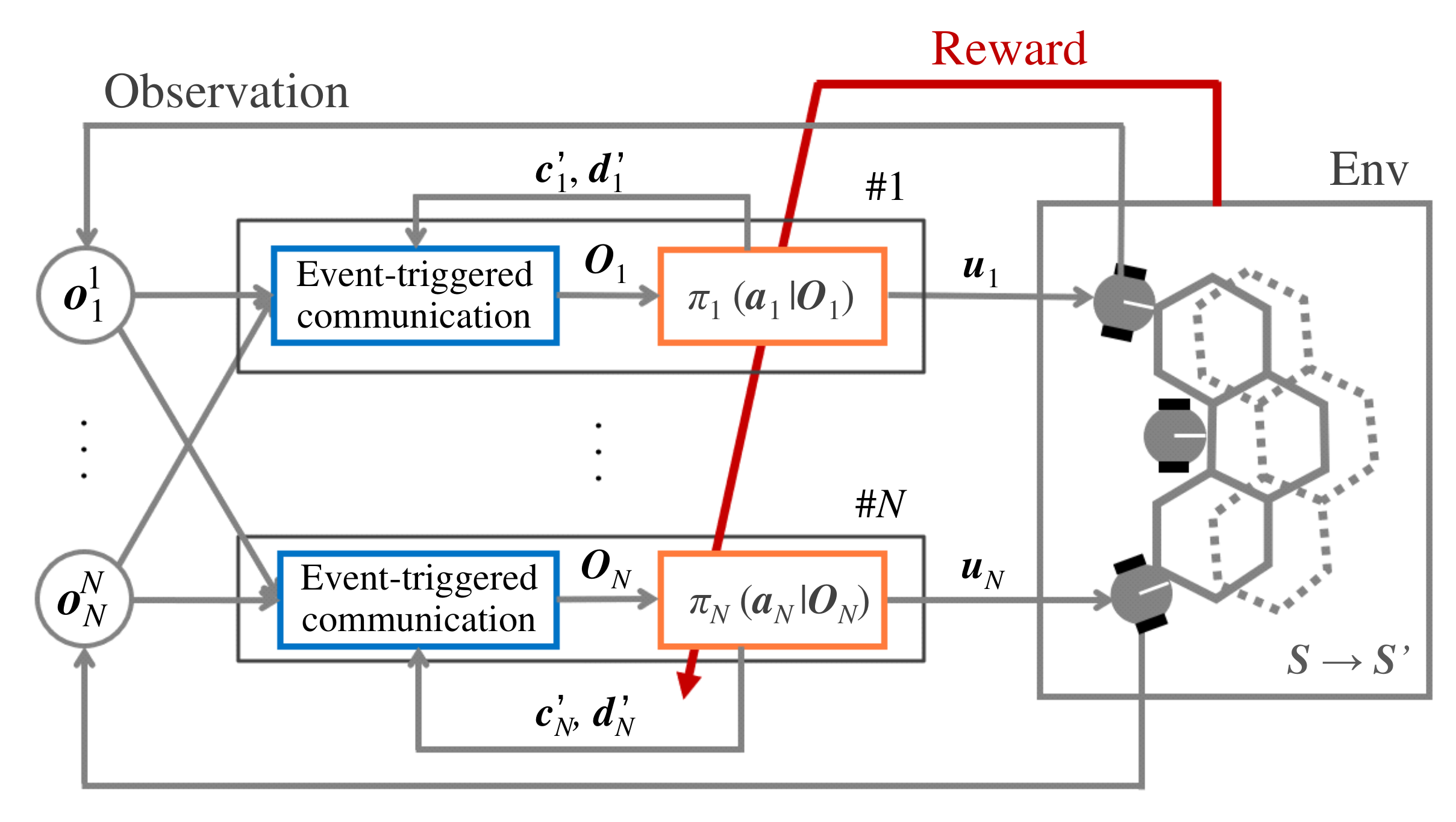}
\label{overview}}
\vspace{-2mm}
\subfigure[Event-triggered communication architecture]{
\includegraphics[width=7.5cm, height=4cm]{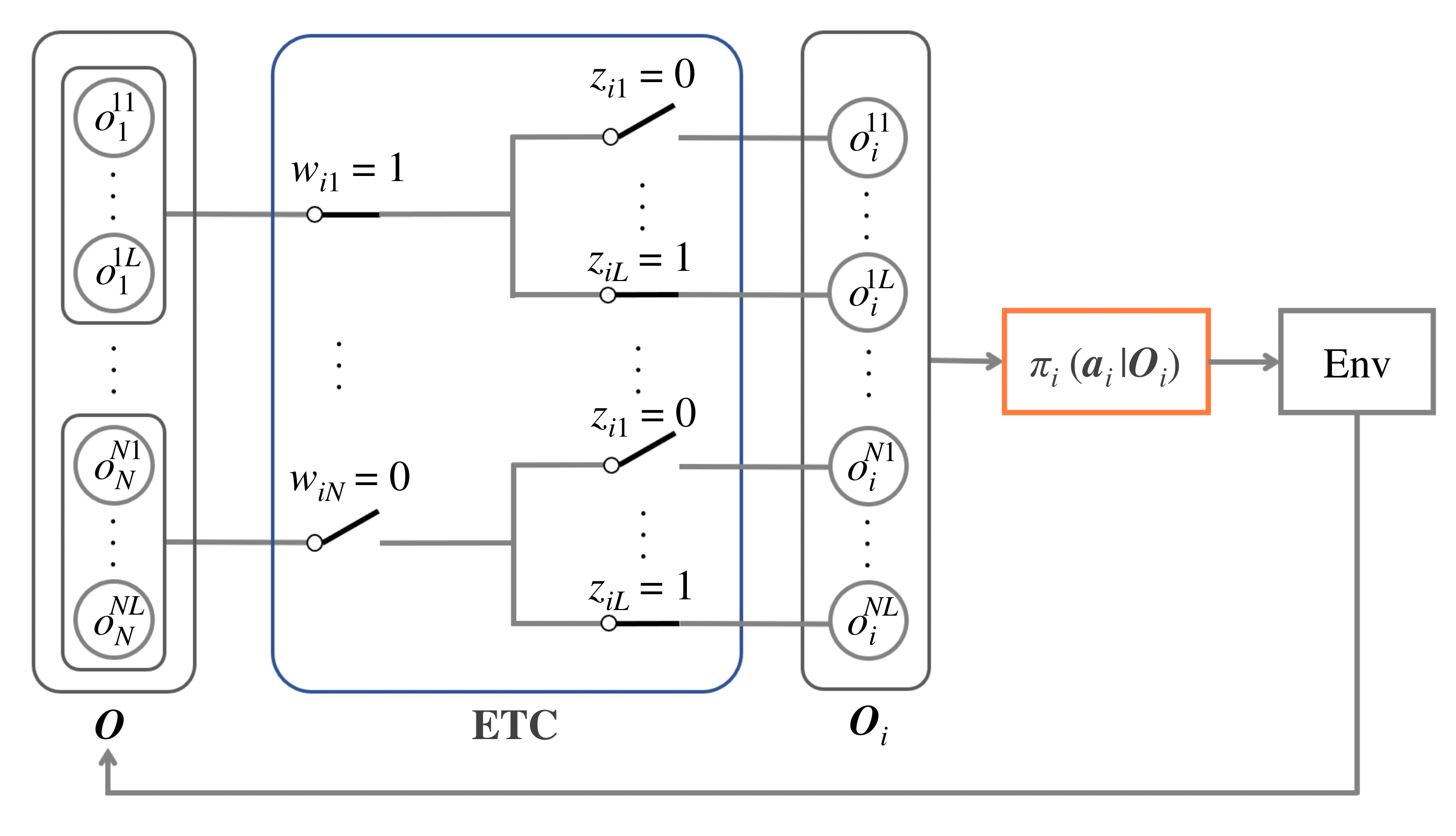}
\label{ETClaw}}
\caption{Our learning framework of event-triggered communication and control}
\label{method}
\end{center}
\end{figure}

Moreover, we designed a joint policy of communication and control given by
\begin{eqnarray}
\pi_i(\textit{\textbf{a}}_i\mid \mathcal{O}_i)=\pi_i(\textit{\textbf{u}}_i, \textit{\textbf{c}}_i, \textit{\textbf{d}}_i \mid \mathcal{O}_i),
\label{policy}
\end{eqnarray}
where $\textit{\textbf{c}}_i=\left[c_{i1},\cdots,c_{iN}\right]^{\rm{T}}\in \mathbb{R}^{N}$ and $\textit{\textbf{d}}_i=\left[d_{i1},\cdots,d_{iL}\right]^{\rm{T}}\in \mathbb{R}^{L}$ represent the communication inputs.
The joint policy in (\ref{policy}) is computed by DNN.
The communication inputs are used to update the observations in the next control step using (2)--(4), which are represented by $\textit{\textbf{c}}'_{i}$ and $\textit{\textbf{d}}'_{i}$ in Fig. \ref{overview}.

\subsection{Reward Design}
To balance the control performance and communication savings, we design the reward given by 
\begin{eqnarray}
r_i=-\| \textit{\textbf{x}}^{\ast}-\textit{\textbf{x}}\|_2-\lambda (\| {\textit{\textbf{w}}_i} \|_1+\| {\textit{\textbf{z}}_i} \|_1),
\label{reward}
\end{eqnarray}
where $\lambda>0$, $\| \bullet \|_1$ and $\| \bullet \|_2$ represent the hyperparameter, $L_1$ and $L_2$ norm, respectively.
The $L_1$ norm used in the second term aims to minimize the number of agents to communicate and the data to be received at every time step.

In the learning process, the weight parameters in the DNN policies are optimized to maximize the total reward.
One of the main problems in MARL is that the variance of the estimated policy becomes large due to the changing policies of other agents in partially observable environments.
Various MARL methods solve this problem, including counterfactual multi-agent policy gradients (COMA) \cite{Foerster2018} and multi-agent deep deterministic policy gradient (MADDPG) \cite{Lowe2017}. Our framework can be combined with these MARL methods.

\section{Simulation}
We conducted several cooperative transport simulations to confirm that our algorithm could balance the transport performance and communication savings.

\subsection{Cooperative Transport with Agents Rigidly Attached to a Payload}
We start with cooperative transport simulations with agents rigidly attached to a payload whose minimum-communication topology is analytically tractable \cite{Culbertson2018}.
Through this problem, we confirm that the payload position can be controlled to be the desired position without communication between agents, using the current and desired positions of the payload broadcasted to all agents.
\subsubsection{Setup}
Let us consider a 2D cooperative transport problem using a team of point agents rigidly attached to the edges of the payload \cite{Culbertson2018}.
The payload shape is a square. The size, mass, and inertia of moment are set to be 0.5 $\times$0.5 $\rm{}m^2$, 1.0 kg, and 4.2$\times10^{-2}$ kg$\rm{}m^2$, respectively.
The control input of agent $i$ is $\textit{\textbf{u}}_i=[f_{xi}, f_{yi}]^{\rm{T}}$, where $f_{xi}$ and $f_{yi}$ represent the force in the $x$ and $y$ axes in world coordinates, respectively.


In this study, we used MADDPG \cite{Lowe2017}, which is a deep actor critic algorithm, to optimize the multi-agent policies.
Numerical simulations were carried out using the code in \cite{MADDPGcode} and the dynamics presented in \cite{WangDARS2016}.
Table I lists the simulation parameters.
The parameters used in the MARL method were set by trial and error.
We carried out 10 trainings under the same conditions.

We set the communication data between different agents to be the positions, velocities, and forces.
The observations, action, and reward of agent $i$ are given by $\textit{\textbf{o}}_i=[\textit{\textbf{b}}^{\rm{T}}\it{}, {\textit{\textbf{x}}_i}^{\rm{T}}\it{}, {\textit{\textbf{v}}_i}^{\rm{T}}, {\textit{\textbf{u}}_i}^{\rm{T}}]^{\rm{T}}$, $\textit{\textbf{a}}_i=[{\textit{\textbf{u}}_i}^{\rm{T}},{{\textit{\textbf{c}}_i}}^{\rm{T}},{{\textit{\textbf{d}}_i}}^{\rm{T}} ]^{\rm{T}}$, $r_i=-\| \textit{\textbf{x}}^{\ast}-\textit{\textbf{x}}\|_2-\lambda (\| {\textit{\textbf{w}}_i} \|_1+\| {\textit{\textbf{z}}_i} \|_1)$, where $\textit{\textbf{b}}=[\textit{\textbf{x}}^{\rm{T}}\it{}, \textit{\textbf{v}}^{\rm{T}}\it{}, \theta, \omega, {\textit{\textbf{x}}^{\ast}}^{\rm{T}}\it{}]^{\rm{T}}$ represents the signal of the payload broadcasted to all agents.
We set $\lambda=0.2$ by trial and error in (\ref{reward}).

\begin{table}[!bp]
\caption{Simulation conditions}
\label{sim1_condition}
\centering
\renewcommand{\arraystretch}{1.1}
\vspace{-2mm}
\begin{tabular}{cc}
\hline
Variable & Value \\
\hline
Control period [s] & 0.1 \\
Time step size of dynamics [s] & 0.1 \\
Number of steps per episode & 25 \\
Number of episode & 2.0$\times10^{5}$ \\
Number of hidden layers (critic) & 4 \\
Number of hidden layers (actor) & 4 \\
Number of units per layer & 64 \\
Activation function of hidden layers & ReLU \\
Activation function of output layers (critic) & linear \\
Activation function of output layers (actor) & tanh \\
Discount factor & 0.99 \\
Batch size & 256 \\
Replay buffer & 1.0$\times10^6$ \\
Translational friction coefficient (payload vs floor) & 0.3 \\
Rotational friction coefficient (payload vs floor) & 1.0$\times10^1$ \\
\hline
\end{tabular}
\end{table}

Moreover, we define the transport performance and communication cost as follows:
\begin{eqnarray}
E&=&-\sum^T_{k=1}\| \textit{\textbf{x}}^{\ast}-\textit{\textbf{x}}(k)\|_2, \label{performance} \\
C&=&\sum^N_{i=1} \sum^T_{k=\rm{1}\it{}} \sum^N_{j=\rm{1}\it{}} \sum^L_{l=\rm{1}\it{}}\zeta^{jl}_{i}(k), \label{comcost}
\end{eqnarray}
where $T$ and $\zeta^{jl}_{i}(k)$ represent the total number of steps during one episode and a binary variable, respectively.
Specifically, $\zeta^{jl}_{i}(k)=1$ if agent $i$ receives the $l$ th element of observations held by agent $j$ at control time step $k$; otherwise, $\zeta^{jl}_{i}(k)=0$.

\subsubsection{Result}
The position of the payload can be controlled to its desired position if each agent can obtain all data from other agents at a high fixed rate.
Therefore, we start to confirm that our method shows as good transport performance as that of high-fixed-rate communication.

\begin{figure}[!bp]
\begin{center}
\vspace{-5mm}
\includegraphics[width=7cm]{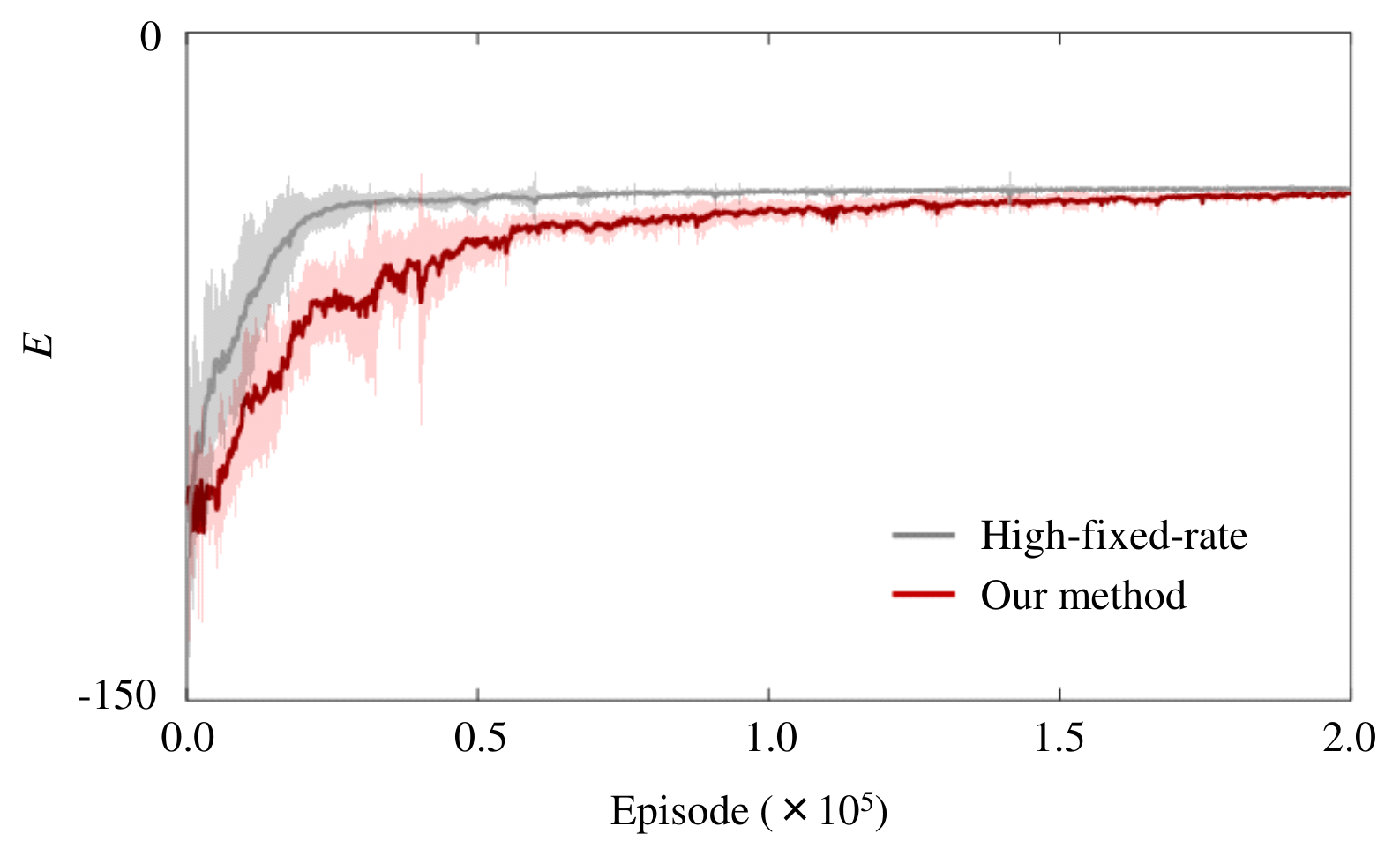}
\caption{Comparisons of the transport performnace}
\label{lcurvesim1}
\end{center}
\end{figure}

\begin{figure}[!bp]
\centering
\vspace{-3mm}
\subfigure[High fixed rate]{
\includegraphics[width=3.5cm]{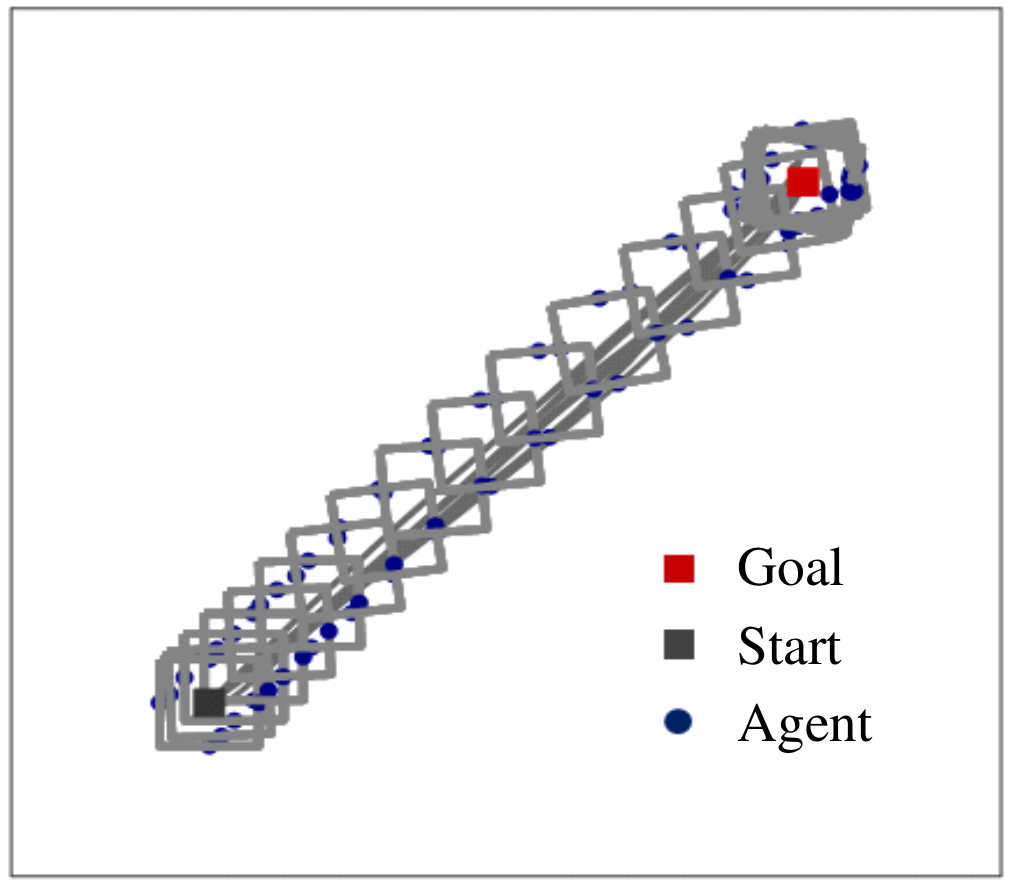}
\label{trakectorymaddpgsim1}}
\subfigure[Our method]{
\includegraphics[width=3.5cm]{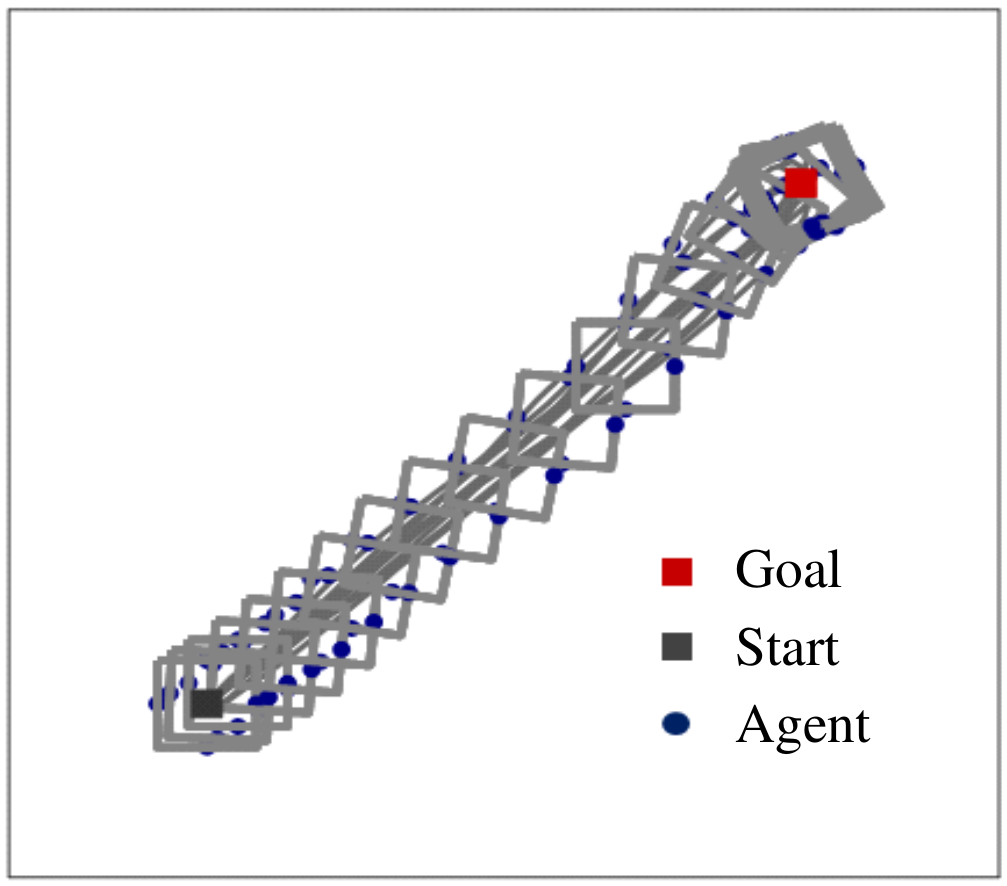}
\label{trakectoryetcsim1}}
\caption{Trajectories of the payload at the end of training}
\label{sim1trajectory}
\end{figure}

The transport performance in (\ref{performance}) at each episode when applying both methods is shown in Fig. \ref{lcurvesim1}.
The results show that our method makes the transport performance converge to almost the same value as that of high-fixed-rate communication.
After each training, the position of the payload could be controlled to a position close to its desired position when applying both methods, as shown in Fig. \ref{sim1trajectory}.

\begin{figure}[!tp]
\begin{center}
\includegraphics[width=7cm]{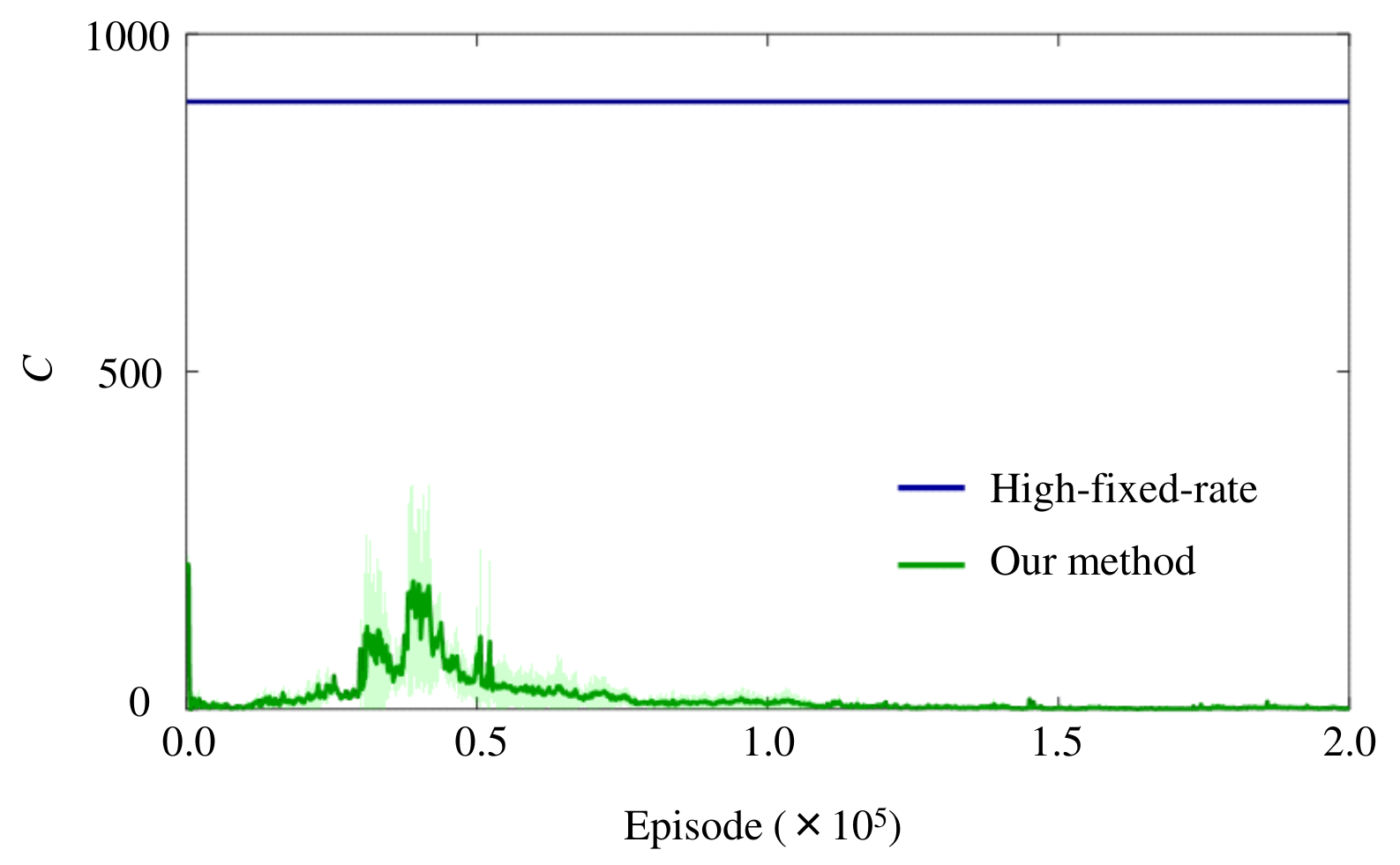}
\caption{Comparisons of the communication cost}\label{comcurvesim1}
\end{center}
\end{figure}

Moreover, the communication cost in (\ref{comcost}) at each episode is shown in Fig. \ref{comcurvesim1}.
The results show that our method makes the communication cost almost equal to zero.
This indicates that each agent never received either positions, velocities, or forces from other agents when applying our method.

In summary, our framework can derive both control policies to achieve the task and the communication-free topology for the cooperative transport problem with a team of agents rigidly attached to the payload, which is consistent with the model-based analytical study \cite{Culbertson2018}.

\subsection{Cooperative Pushing with Actuator Failure}
In this subsection, we confirm the versatility of our framework through cooperative pushing with actuator failure, whose minimum-communication topology is analytically intractable.
\subsubsection{Setup}
Let us consider a 2D cooperative pushing, in which a certain agent stops owing to actuator failure.
The mass and moment of inertia are set to be 4.8 kg and 1.3$\rm{}\times10^{-2}$ kg$\rm{}m^2$, respectively.
The shape of the agent is circular. The radius, mass, and moment of inertia of the agent are set to be 0.10 m, 1.1 kg, and 5.3$\rm{}\times10^{-3}$ kg$\rm{}m^2$, respectively.
The control input of agent $i$ is $\textit{\textbf{u}}_i=[u^v_i, u^{\omega}_i]^{\rm{T}}$, where $u^v_i\in \mathbb{R}$ and $u^{\omega}_i\in \mathbb{R}$ represent the velocity input in the forward direction and the angular velocity, respectively.
We set $\mid u^v_i\mid \le 0.2$ and $\mid u^{\omega}_i\mid \le 0.5$, considering the robot used in the experiment.
We make agent 1 stop after 5 [s] by setting $u^v_1=0.0$ and $u^{\omega}_1=0.0$.
Note that each agent cannot directly know the time of failure and its position, which changes at every episode.

The parameters used in the simulations are listed in Table II.
The numbers of layers and units, activation functions, and discount factors are omitted because they are the same as those shown in Table I. 
In the simulations, we integrated the spring mass models to the dynamics of the payload and agents presented in \cite{WangDARS2016}.
We carried out three trainings by randomly setting the initial yaw angle of the payload $-\pi/8\le \theta \le \pi/8$ while fixing the initial position of the payload and agents. 

Moreover, we set the communication data to be the positions, yaw angles, velocities, angular velocities, and control inputs.
The observations, action, and reward are given by $\textit{\textbf{o}}_i=[\textit{\textbf{b}}^{\rm{T}}\it{}, {\textit{\textbf{x}}_i^B}^{\rm{T}}, {\textit{\textbf{v}}_i^B}^{\rm{T}}, \theta_i^B, \omega_i^B, {\textit{\textbf{u}}_i^{\rm{T}}\it{}}]^{\rm{T}}$, $\textit{\textbf{a}}_i=[{\textit{\textbf{u}}_i}^{\rm{T}}\it{},{\textit{\textbf{c}}_i}^{\rm{T}}\it{},{{\textit{\textbf{d}}_i}}^{\rm{T}} ]^{\rm{T}}$, $r_i=-\| \textit{\textbf{x}}^{\ast}-\textit{\textbf{x}}\|_2-\lambda (\| \textit{\textbf{w}}_i \|_1+\| \textit{\textbf{z}}_i \|_1)$, 
where the superscripts $B$ and $\textit{\textbf{b}}=[\textit{\textbf{x}}^{\rm{T}}\it{}, \textit{\textbf{v}}^{\rm{T}}\it{}, \theta, \omega, {{\textit{\textbf{x}}^{\ast}}^B}^{\rm{T}}\it{}]^{\rm{T}}$ represent the values in the payload coordinate, and the signal of the payload is broadcasted to all agents.
Moreover, we set $\lambda=0.01$ by trial and error in (\ref{reward}).


\begin{table}[!bp]
\caption{Simulation conditions}
\label{sim2_condition}
\vspace{-2mm}
\centering
\renewcommand{\arraystretch}{1.1}
\begin{tabular}{cc}
\hline
Variable & Value \\
\hline
Control period [s] & 0.25 \\
Time step size of dynamics [s] & 0.05 \\ 
Number of steps per episode & 150 \\
Number of episode & 2.0$\times10^{5}$ \\
Batch size & 4096 \\
Replay buffer & 2.0$\times10^6$ \\
Translational friction coefficient (payload vs floor) & 0.2 \\
Rotational friction coefficient (payload vs floor) & 1.5$\times10^{1}$ \\
Spring constant (payload vs agent) & 3.0$\times10^{2}$ \\
Spring constant (agent vs agent) & 2.0$\times10^{1}$ \\
\hline
\end{tabular}
\end{table}

\subsubsection{Result}
To confirm the effectiveness of our algorithm, we compared our algorithm with several communication topologies as follows.
\begin{itemize}
\item High-fixed-rate communication: each agent receives all data at every 0.25 s, which is the same as the control period.
\item No communication: each agent receives no data from the other agents throughout one episode. 
\item Low-fixed-rate communication: each agent receives all data at every 5 s, which is higher than the rate at which the communication cost is the same as that in our method.
\end{itemize}

\begin{figure}[!bp]
\begin{center}
\vspace{-3mm}
\includegraphics[width=8cm]{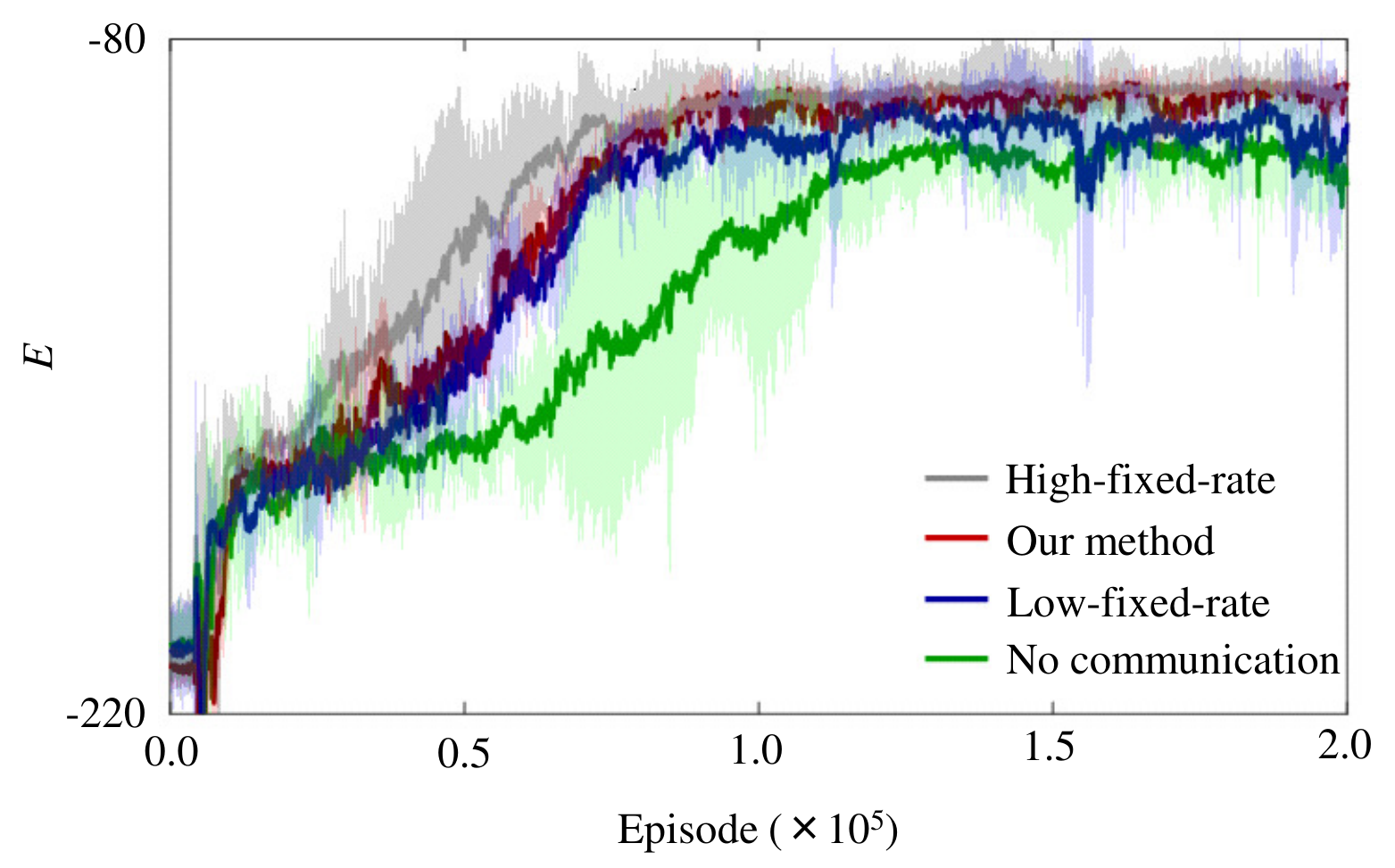}
\caption{Comparison of the transport performance}\label{lcurvesim2}
\end{center}
\end{figure}

Fig. \ref{lcurvesim2} compares the transport performance in (\ref{performance}) at each episode.
The results show that our method makes the transport performance converge to almost the same value as that of high-fixed-rate communication.
To compare the communication topologies quantitatively, we execute the trained policies 1000 times and evaluate the success rate at which the payload position at the last step is within 0.2 m from the desired position. According to the result, our method achieves a success rate as high as that of the high-fixed-rate communication, as summarized in Table \ref{rate}.
At the same time, our method makes the communication cost smaller than the high- and low-fixed-rate communication, as shown in Table \ref{comamounttable}.

\begin{table}[!bp]
\caption{Comparison of success rate}
\vspace{-3mm}
\centering
\renewcommand{\arraystretch}{1.2}
\begin{tabular}{cccc}
\hline
High-fixed-rate & Low-fixed-rate & No communication & Our method \\
\hline
0.99 & 0.87 & 0.84 & \textbf{0.99} \\
\hline
\label{rate}
\end{tabular}
\end{table}

\begin{table}[!tp]
\caption{Comparison of communication cost}
\vspace{-3mm}
\centering
\renewcommand{\arraystretch}{1.2}
\begin{tabular}{ccc}
\hline
High-fixed-rate & Low-fixed-rate & Our method \\
\hline
8.1$\times10^3$ & 4.1$\times10^2$ &  \textbf{1.3$\times10^2$} \\
\hline
\label{comamounttable}
\end{tabular}
\end{table}

\begin{figure}[!tp]
\centering
\subfigure[Communication topologies]{
\includegraphics[width=6.8cm]{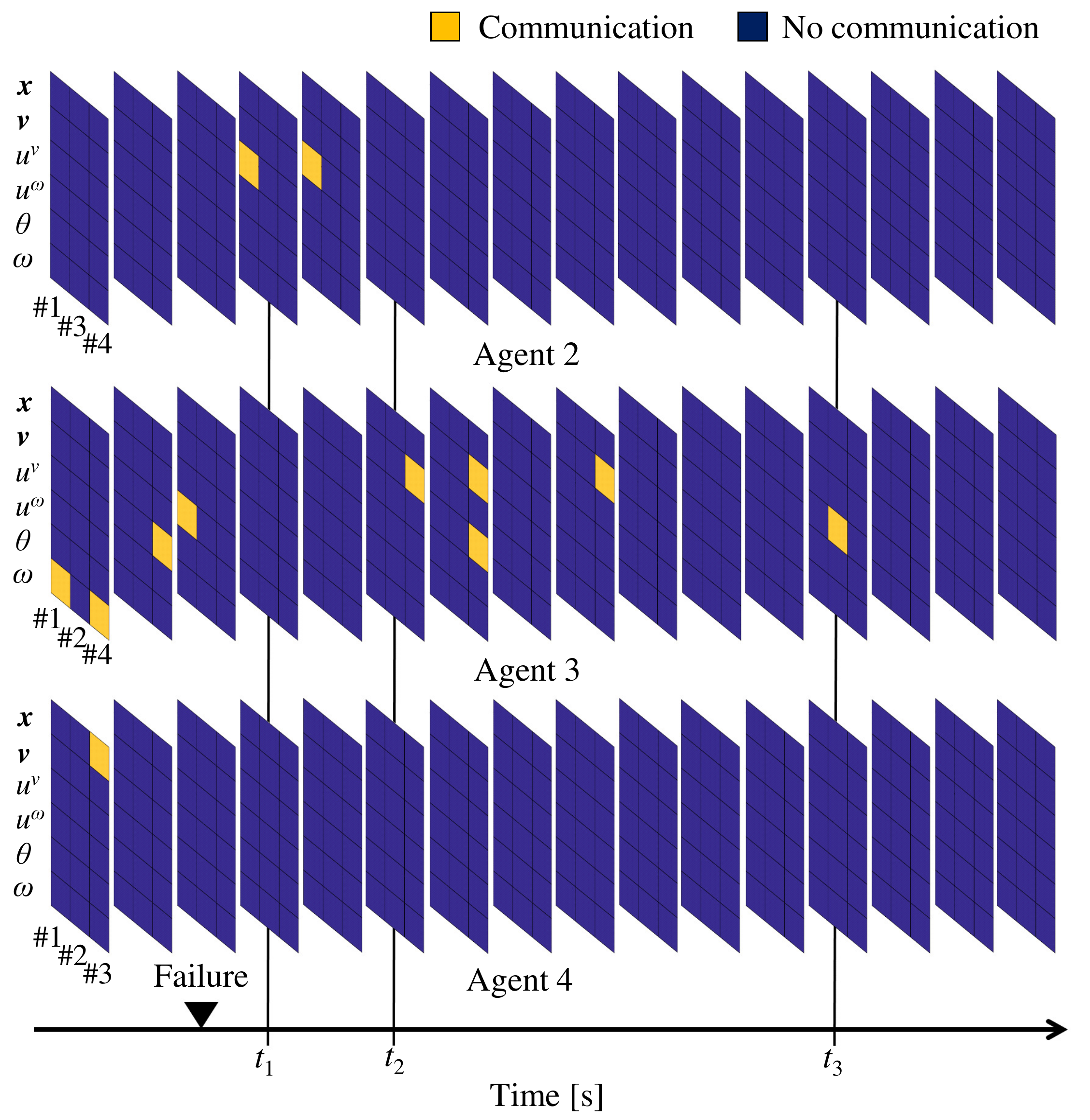}
\label{com}}
\subfigure[States]{
\includegraphics[width=8cm, height=2.5cm]{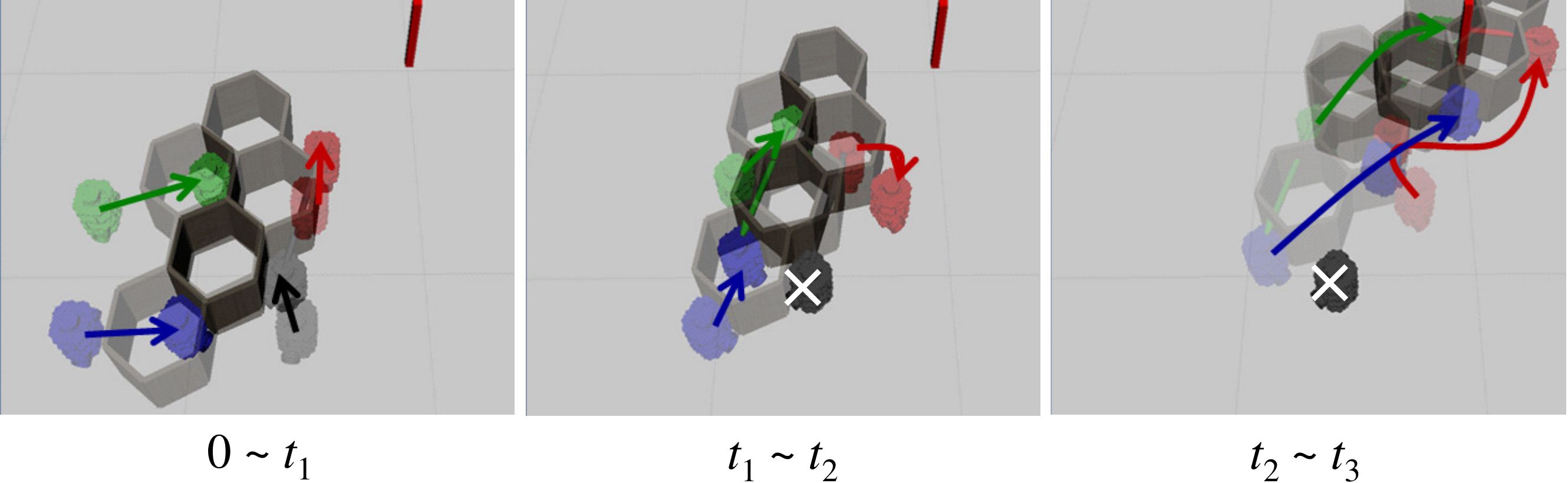}
\label{state}}
\caption{Communication topologies and states at the end of training when applying our algorithm.
The colored grid of each cross section represents whether each agent receives $\textit{\textbf{x}}$, $\textit{\textbf{v}}$, $u_v$, $u_{\omega}$, $\theta$, and $\omega$ from other agents or not, at every 2.5 s.}
\label{comsim2}
\end{figure}

The communication topologies and states obtained by our algorithm after training are shown in Fig. \ref{comsim2}.
Just after the actuator failure of agent 1, agent 2 received the velocity input from agent 1 and changed its pushing position drastically.
This result indicates that agent 2 knew the occurrence of the failure based on the velocity input and thereby selected its action. Moreover, at the end of the task, agent 3 received the velocity input, whereas it tried to make the position of the payload converge to the desired position.  

In summary, our framework can achieve more communication savings than other communication topologies while maintaining as good transport performance as that of high-fixed-rate communication, even for a complex task whose minimum-communication topology is analytically intractable.

\section{Demonstration}
\begin{figure}[!tp]
\begin{center}
\includegraphics[width=4.5cm]{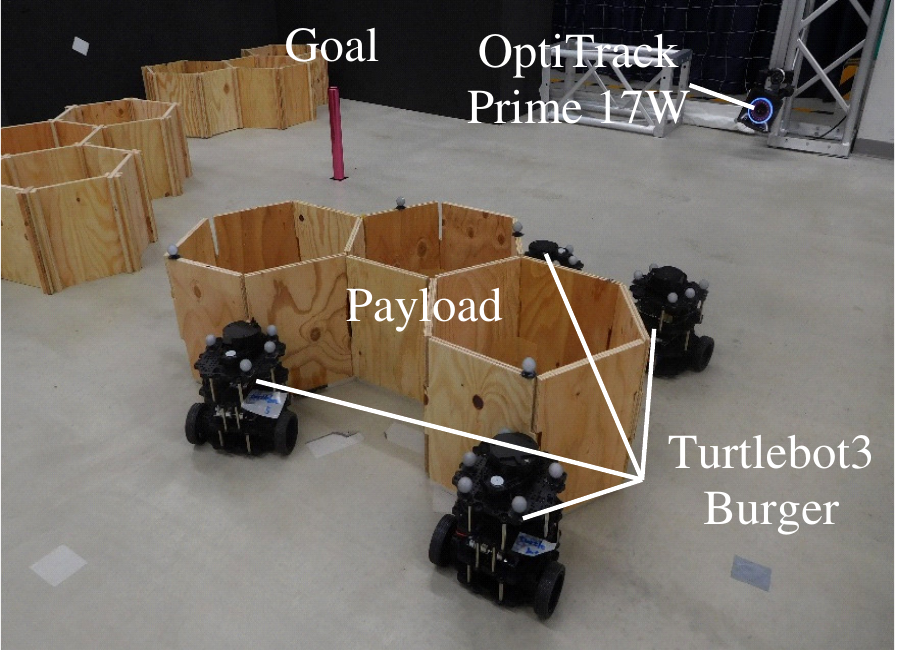}
\caption{Experimental configuration}
\label{config}
\end{center}
\end{figure}

The experimental configuration is shown in Fig. \ref{config}. The aim of this demonstration is to confirm that the trained policies in the simulation can control the payload to its desired position using multiple real robots.

Our framework may be used in several communication scenarios. 
In a client-server scenario, each agent can determine the agents and data to receive from a server which aggregates observations obtained from the infrastructure (e.g., motion capture system).
In a peer-to-peer scenario, each agent requests the necessary data from agents to communicate.
In this study, we performed a demonstration using the client--server scenario.

The positions and yaw angles of the payload and the agents were observed used motion capture system operating at 120 Hz.
The velocities and angular velocities were calculated using the measured positions and yaw angles.
The control inputs were calculated in a control PC with an 8-core Intel$^{\textregistered}$ Core$^{\texttrademark}$ i7 (2.80 GHz), 32 GB of RAM, using the policies learned in the simulations from these measurements and sent to each robot at 4 Hz via Wi-Fi communication.

\begin{figure}[!tp]
\begin{center}
\vspace{-3mm}
\includegraphics[width=7.5cm]{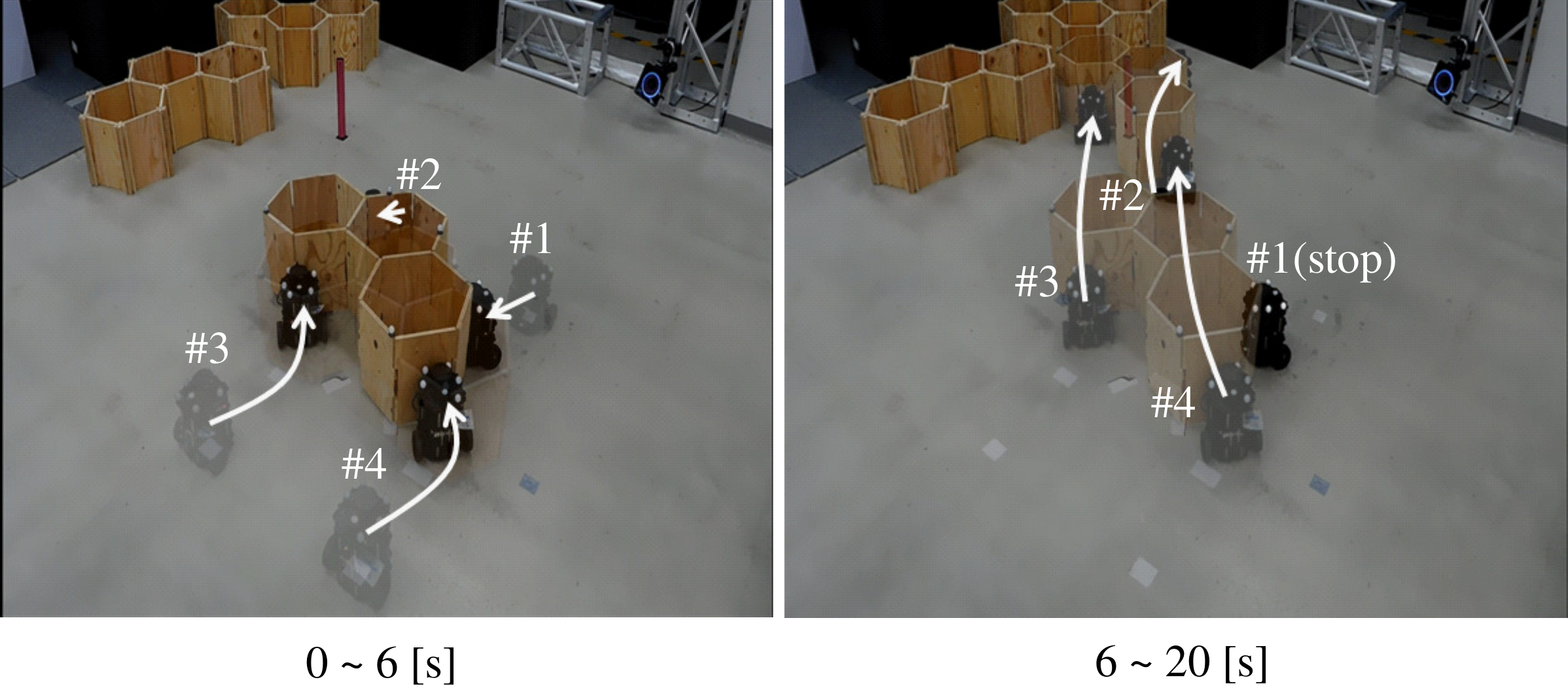}
\caption{Real robot demonstration}
\label{demo}
\end{center}
\end{figure}

The results of the real robot demonstrations are shown in Fig. \ref{demo}.
Despite gaps between simulations and real experiments, the payload could be controlled to a position close to its desired position for several trials when applying multi-agent policies learned in the simulations.

\section{Discussion}
In future work, we plan to apply our algorithm to system design for cooperative manipulation.
Let us consider a cooperative transport problem with the minimum required communication between the agents and its mounted sensors.
By applying our algorithm, we could derive the control policy with the minimum required observations and determine the minimum sensor configuration for cooperative manipulation.


The trained multi-agent policies performed worse in the real experiments than in the simulations because the physical parameters of the simulations are different from those of the experiments. To address this issue, we plan to make multi-agent policies more robust using domain randomization \cite{Tobin} and evaluate the trained policies quantitatively.

Moreover, we should examine how our algorithm works under partially observable environments using the recurrent multi-agent
deep deterministic policy gradient model presented in \cite{REWang}.

\section{Concluson}
In this paper, we proposed a learning framework that balanced the transport performance and communication savings.
We confirmed that, for cooperative transport with agents rigidly attached to a payload, our approach could derive the communication-free topology while maintaining as good transport performance as that of high-fixed-rate communication.
Moreover, for cooperative pushing with actuator failure, our approach could achieve more communication savings than other communication topologies while maintaining as good transport performance as that of high-fixed-rate communication.
In future work, we plan to apply our approach to system design for cooperative manipulation.


\end{document}